\documentclass[runningheads]{llncs}
\usepackage[T1]{fontenc}
\usepackage{graphicx}
\PassOptionsToPackage{table, dvipsnames, svgnames, x11names}{xcolor}

\usepackage{colortbl}       
\usepackage{booktabs}
\usepackage{multirow}
\usepackage{tikz}
\usetikzlibrary{shadows}
\usetikzlibrary{shadows.blur, backgrounds}
\usepackage{booktabs}
\usepackage{multirow}
\usepackage{tcolorbox}
    \tcbuselibrary{skins}
\newtcolorbox{mybox}[1]{
    colback=brown!5!white,
    colframe=brown!5!black,
    borderline={2pt}{0mm}{black},
    borderline={.7pt}{1mm}{black},
    fonttitle=\bfseries,
    title=#1,
    arc=3mm,
    segmentation hidden,
    top=8pt,
    bottom=8pt
}

\usepackage{latexsym}
\usepackage{enumitem}
\usepackage{graphicx}

\usepackage[
  colorlinks=true,
  linkcolor=blue,
  citecolor=blue,
  urlcolor=blue
]{hyperref}

\usepackage{orcidlink}
\usepackage{url}

\begin{document}
%
\title{Grounding AI-in-Education Development in Teachers’ Voices: Findings from a National Survey in Indonesia}
\titlerunning{Teachers’ Voices: Findings from a National Survey in Indonesia}
\author{Nurul Aisyah\inst{1}\orcidlink{0009-0003-3500-831X} \and
Muhammad Dehan Al Kautsar\inst{2}\orcidlink{0009-0000-5416-3524} \and \\
Arif Hidayat\inst{3}\orcidlink{0000-0003-0734-0756} \and
Fajri Koto\inst{2}\orcidlink{0000-0002-3659-6761}
}

\authorrunning{N. Aisyah et al.}
%
\institute{Quantic School of Business and Technology, Washington, DC, USA \\ \email{na57@students.quantic.edu}
 \and
Mohamed bin Zayed University of Artificial Intelligence, Abu Dhabi, UAE 
\\ \email{muhammad.dehan@mbzuai.ac.ae, fajri.koto@mbzuai.ac.ae}
\and
Indonesia University of Education, Bandung, Indonesia 
\\ \email{arifhidayat@upi.edu}
}

\maketitle              
\begin{abstract}


Despite emerging use in Indonesian classrooms, there is limited large-scale, teacher-centred evidence on how AI is used in practice and what support teachers need, hindering the development of context-appropriate AI systems and policies. To address this gap, we conduct a nationwide survey of 349 K–12 teachers across elementary, junior high, and senior high schools. We find increasing use of AI for pedagogy, content development, and teaching media, although adoption remains uneven. Elementary teachers report more consistent use, while senior high teachers engage less; mid-career teachers assign higher importance to AI, and teachers in Eastern Indonesia perceive greater value. Across levels, teachers primarily use AI to reduce instructional preparation workload (e.g., assessment, lesson planning, and material development). However, generic outputs, infrastructure constraints, and limited contextual alignment continue to hinder effective classroom integration.


\end{abstract}
\section{Introduction}

Indonesia’s education system serves more than 52 million students but relies on only around three million teachers,\footnote{\small \url{https://www.statista.com/topics/9229/education-in-indonesia}} making scalable approaches to improving learning outcomes essential. Although artificial intelligence (AI) is increasingly being introduced into Indonesian classrooms, large-scale, teacher-centred evidence on how AI is adopted in practice and what kinds of support teachers require remains limited. This lack of understanding hampers the development of AI systems that align with classroom realities, particularly in a context marked by uneven digital infrastructure, varying technological readiness, and linguistic diversity \cite{kusharjanto2011infrastructure,rahmi2024challenges,harsanti2025exploring,unesco2024airamindonesia}.

In this paper, we present findings from a nationwide survey of Indonesian K–12 teachers examining their adoption of AI tools across six key domains of instructional practice: content knowledge development, pedagogy \cite{kleickmann2013teachers}, assessment and evaluation \cite{aisyah2025evaluating}, curriculum development \cite{alsubaie2016curriculum}, teaching media, and professional practice. Beyond documenting usage patterns, we investigate teachers’ perceived needs for AI-based support and the challenges they face when integrating AI into their work. Following the taxonomy of Al-Kautsar et al. \cite{al-kautsar-etal-2025-indonesians}, we categorize AI tools into six groups: LLM-based text assistants, machine translation systems, image generation tools, video generation tools, grammatical correction tools, and speech technologies, including both text-to-speech and speech-to-text.

This study makes two primary contributions. First, it provides a national map of AI-related needs in Indonesian education, disaggregated by key teacher and school characteristics. Second, it offers evidence-based recommendations for educational technology developers and AI-in-education initiatives.


\section{Related Work}


Prior work by Diliberti et al. \cite{diliberti2024using} examined AI use in teaching within U.S. schools but remained limited in its coverage of usage frequency across educational levels. Other studies, such as Wen et al. \cite{wen2024chatgptteachers}, surveyed teachers in Brazil, Israel, Japan, Norway, Sweden, and the United States to understand factors influencing teachers’ willingness to use AI in education. Our work differs in both focus and depth. We study AI adoption in Indonesia, a developing-country context, and examine how AI is used across specific components of teaching practice, as well as the concrete challenges teachers face and the support they need in everyday classroom settings.

Prior work on AI use in Indonesian education exists but remains limited in scope. Widianingsih \cite{widianingsih2025transforming} examines changes in teachers’ roles and training needs following AI adoption, while Putra et al. \cite{putra2025utilization} focus narrowly on AI use in \textit{Bahasa Indonesia} instruction based on interviews with only twelve teachers. Other studies emphasize systemic issues—such as infrastructure, policy, privacy, and multilingualism—rather than everyday classroom practice \cite{maspul2025can,fauziddin2025impact}. In contrast, our work draws on a nationwide survey to examine how Indonesian teachers use AI across core teaching components and across a broad range of teacher characteristics, including age, subject area, and education level, while identifying concrete challenges and support needs.

\section{Methods}

We conducted a nationwide, cross sectional online survey to examine Indonesian K–12 teachers’ use of, needs for, and concerns about AI across key instructional functions. Using non probability recruitment through education offices, teacher networks, and digital channels, we collected 349 valid responses from teachers across at least 25 provinces representing all major islands in Indonesia, covering diverse school levels, subjects, and urban and rural contexts.

The survey was developed in Indonesian by a multidisciplinary focus group consisting of two education experts holding PhD and MSc degrees and one researcher with a PhD in artificial intelligence. It examined teachers’ use of AI across six components of teaching and learning, namely content knowledge development, pedagogy, assessment and evaluation, curriculum development, teaching media, and professional practice, as well as five types of AI tools, including LLM based text assistants, machine translation, generative image and video tools, grammar checkers, and speech technologies. The survey also captured teachers’ satisfaction with existing tools, their priority needs, and the challenges they experienced when using AI, together with background information such as age, location, subject taught, and teaching experience. Quality control was ensured through the inclusion of attention check questions.

Data were collected over four weeks in 2025 using a low bandwidth compatible platform with voluntary and anonymous participation. Likert scale responses were rescaled to a 0–1 range, and subgroup means were compared as percentage deviations from overall component means. Significance testing was conducted for pairwise comparisons with a threshold of $p < 0.05$.


\begin{figure}[t]
  \centering
  \includegraphics[width=\linewidth]{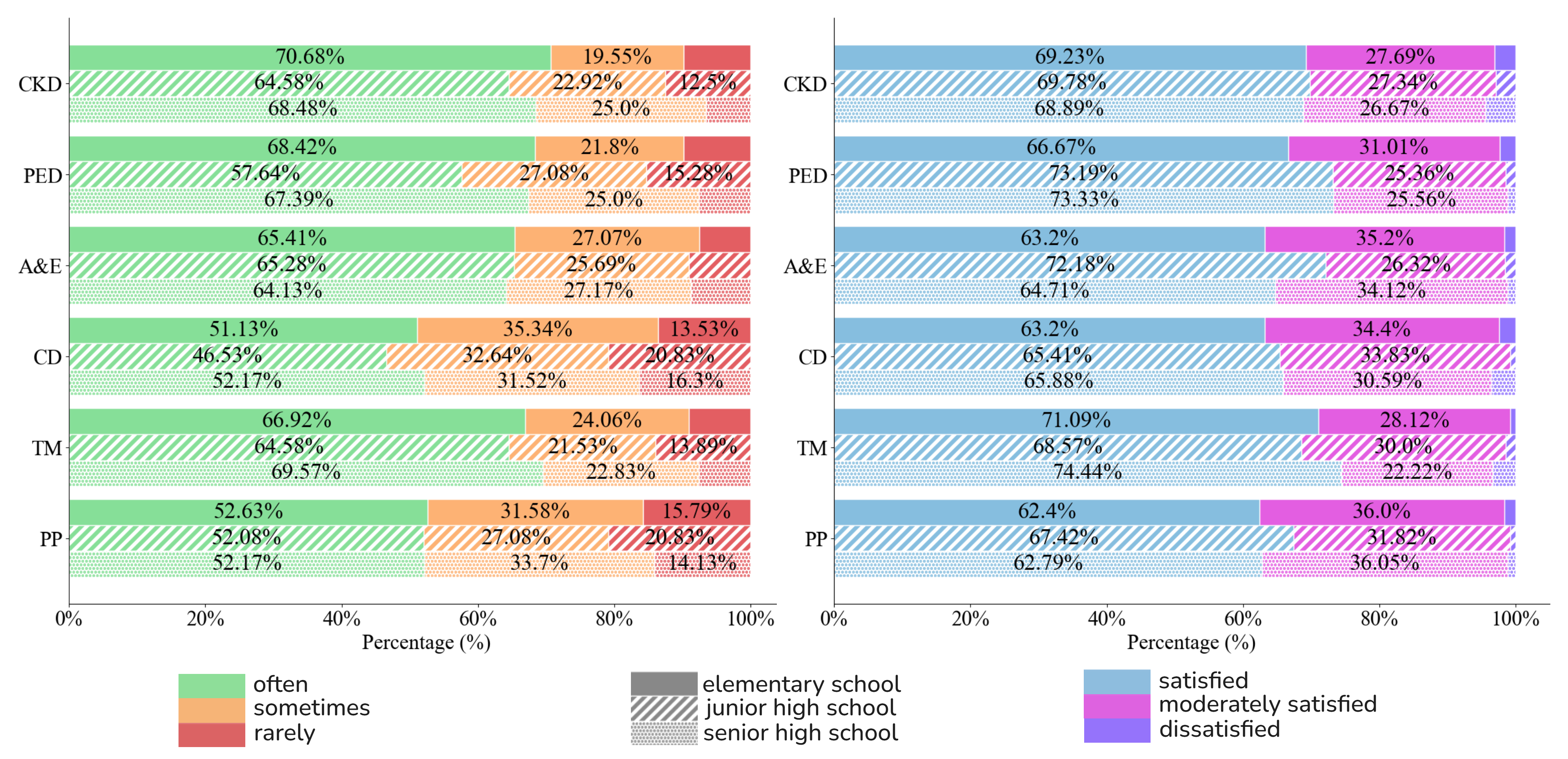}
  \caption {Frequency (left) and satisfaction (right) levels of AI tool usage in Indonesian education across components of effective teaching and learning. We divide the distribution by teachers’ school levels (elementary, junior high, and senior high). CKD = Content Knowledge Development; PED = Pedagogy; A\&E = Assessment and Evaluation; CD = Curriculum Development; TM = Teaching Media; PP = Professional Practice.}
  \label{fig:freq_and_satis_percentage}
\end{figure}

\begin{table}[t]
    \caption{Percentage differences across categories relative to the average importance score for each component of effective teaching and learning. CKD = Content Knowledge Development; PED = Pedagogy; A\&E = Assessment and Evaluation; CD = Curriculum Development; TM = Teaching Media; PP = Professional Practice.}
    \centering
    \small
    \setlength{\tabcolsep}{4pt}
    \renewcommand{\arraystretch}{0.9}
    \resizebox{0.9\linewidth}{!}{%
    \begin{tabular}{l|c|c|c|c|c|c|c}
    \hline
    \textbf{Categories} & \# & \textbf{CKD} & \textbf{PED} & \textbf{A\&E} & \textbf{CD} & \textbf{TM} & \textbf{PP} \\
    \hline
    Full & 349 & \textbf{0.788} & \textbf{0.765} & \textbf{0.785} & \textbf{0.666} & \textbf{0.777} & \textbf{0.673} \\
    \hline
    Science & 100 & \cellcolor{red!15}-1.65\% & \cellcolor{gray!30}0.00 & \cellcolor{red!30}-3.18\% & \cellcolor{red!30}-5.41\% & \cellcolor{blue!15}1.03\% & \cellcolor{blue!30}3.27\% \\
    Social Science & 50 & \cellcolor{red!30}-3.55\% & \cellcolor{red!30}-4.58\% & \cellcolor{red!15}-0.64\% & \cellcolor{red!30}-3.90\% & \cellcolor{red!15}-2.19\% & \cellcolor{blue!15}1.04\% \\
    Language & 75 & \cellcolor{red!15}-1.90\% & \cellcolor{red!15}-2.35\% & \cellcolor{red!30}-4.08\% & \cellcolor{red!30}-6.91\% & \cellcolor{red!30}-4.76\% & \cellcolor{red!15}-0.89\% \\
    General Teacher & 80 & \cellcolor{red!30}-4.06\% & \cellcolor{red!15}-0.39\% & \cellcolor{red!15}-2.04\% & \cellcolor{blue!15}1.35\% & \cellcolor{red!30}-3.47\% & \cellcolor{red!30}-5.20\% \\
    Art and Sport & 24 & \cellcolor{blue!30}5.71\% & \cellcolor{blue!30}6.14\% & \cellcolor{blue!15}0.89\% & \cellcolor{blue!15}0.15\% & \cellcolor{red!30}-3.47\% & \cellcolor{blue!30}5.20\% \\
    Others & 88 & \cellcolor{blue!15}2.41\% & \cellcolor{blue!15}1.05\% & \cellcolor{blue!15}1.27\% & \cellcolor{blue!30}7.51\% & \cellcolor{blue!15}1.67\% & \cellcolor{blue!30}3.86\% \\
    \hline
    Elementary & 133 & \cellcolor{blue!15}2.16\% & \cellcolor{blue!30}3.66\% & \cellcolor{blue!15}0.51\% & \cellcolor{blue!30}3.30\% & \cellcolor{blue!15}1.54\% & \cellcolor{blue!15}1.63\% \\
    Junior High & 144 & \cellcolor{red!30}-3.55\% & \cellcolor{red!30}-6.93\% & \cellcolor{red!15}-0.51\% & \cellcolor{red!30}-5.71\% & \cellcolor{red!30}-3.09\% & \cellcolor{red!30}-2.53\% \\
    Senior High & 192 & \cellcolor{blue!30}2.79\% & \cellcolor{blue!30}4.44\% & \cellcolor{red!15}-1.02\% & \cellcolor{blue!15}1.95\% & \cellcolor{blue!30}4.25\% & \cellcolor{blue!30}2.53\% \\
    \hline
    0-5 Years Experience & 75 & \cellcolor{red!30}-4.44\% & \cellcolor{red!30}-4.97\% & \cellcolor{blue!15}0.25\% & \cellcolor{red!30}-5.86\% & \cellcolor{red!15}-2.19\% & \cellcolor{red!50}-10.85\% \\
    6-10 Years Experience & 89 & \cellcolor{blue!30}9.14\% & \cellcolor{blue!30}4.31\% & \cellcolor{blue!30}8.03\% & \cellcolor{blue!30}6.31\% & \cellcolor{blue!30}7.72\% & \cellcolor{blue!15}1.78\% \\
    11-15 Years Experience & 61 & \cellcolor{blue!50}10.28\% & \cellcolor{blue!30}8.24\% & \cellcolor{red!30}-3.95\% & \cellcolor{blue!30}8.26\% & \cellcolor{blue!30}4.38\% & \cellcolor{blue!30}9.66\% \\
    >15 Years Experience & 124 & \cellcolor{red!30}-8.88\% & \cellcolor{red!30}-4.05\% & \cellcolor{red!30}-3.95\% & \cellcolor{red!30}-4.95\% & \cellcolor{red!30}-6.56\% & \cellcolor{blue!15}0.59\% \\
    \hline
    Public School & 220 & \cellcolor{blue!15}1.27\% & \cellcolor{blue!15}1.31\% & \cellcolor{red!15}-0.13\% & \cellcolor{blue!15}0.30\% & \cellcolor{blue!15}0.64\% & \cellcolor{blue!15}1.04\% \\
    Private School & 135 & \cellcolor{red!15}-1.78\% & \cellcolor{red!15}-1.70\% & \cellcolor{blue!15}0.51\% & \cellcolor{blue!15}0.60\% & \cellcolor{red!15}-0.39\% & \cellcolor{red!15}-0.45\% \\
    \hline
    Undergraduate & 283 & \cellcolor{blue!15}0.51\% & \cellcolor{blue!15}1.57\% & \cellcolor{blue!15}0.89\% & \cellcolor{blue!15}1.35\% & \cellcolor{blue!15}0.90\% & \cellcolor{red!15}-0.30\% \\
    Graduate & 58 & \cellcolor{red!30}-3.68\% & \cellcolor{red!30}-9.80\% & \cellcolor{red!30}-5.61\% & \cellcolor{red!30}-5.56\% & \cellcolor{red!30}-4.63\% & \cellcolor{blue!15}1.19\% \\
    \hline
    West Indonesia & 317 & \cellcolor{red!15}-0.89\% & \cellcolor{red!15}-1.44\% & \cellcolor{red!15}-0.51\% & \cellcolor{red!15}-1.05\% & \cellcolor{red!15}-0.39\% & \cellcolor{red!15}-1.34\% \\
    East Indonesia & 32 & \cellcolor{blue!30}9.01\% & \cellcolor{blue!50}14.38\% & \cellcolor{blue!30}5.48\% & \cellcolor{blue!50}10.21\% & \cellcolor{blue!30}2.57\% & \cellcolor{blue!50}13.82\% \\
    \hline
    Urban & 154 & \cellcolor{red!30}-3.55\% & \cellcolor{blue!15}0.52\% & \cellcolor{red!15}-1.53\% & \cellcolor{blue!15}1.35\% & \cellcolor{red!30}-4.25\% & \cellcolor{blue!15}0.89\% \\
    Suburban & 89 & \cellcolor{blue!30}4.82\% & \cellcolor{blue!15}2.09\% & \cellcolor{blue!15}0.89\% & \cellcolor{blue!15}2.10\% & \cellcolor{blue!30}6.31\% & \cellcolor{red!15}-2.38\% \\
    Village & 106 & \cellcolor{blue!15}1.14\% & \cellcolor{red!30}-2.61\% & \cellcolor{blue!15}1.53\% & \cellcolor{red!30}-3.60\% & \cellcolor{blue!15}0.77\% & \cellcolor{blue!15}0.89\% \\
    \hline
    \end{tabular}}
    \label{tab:assessment_detail}
\end{table}


\begin{figure}[t]
  \centering
  \includegraphics[width=\linewidth]{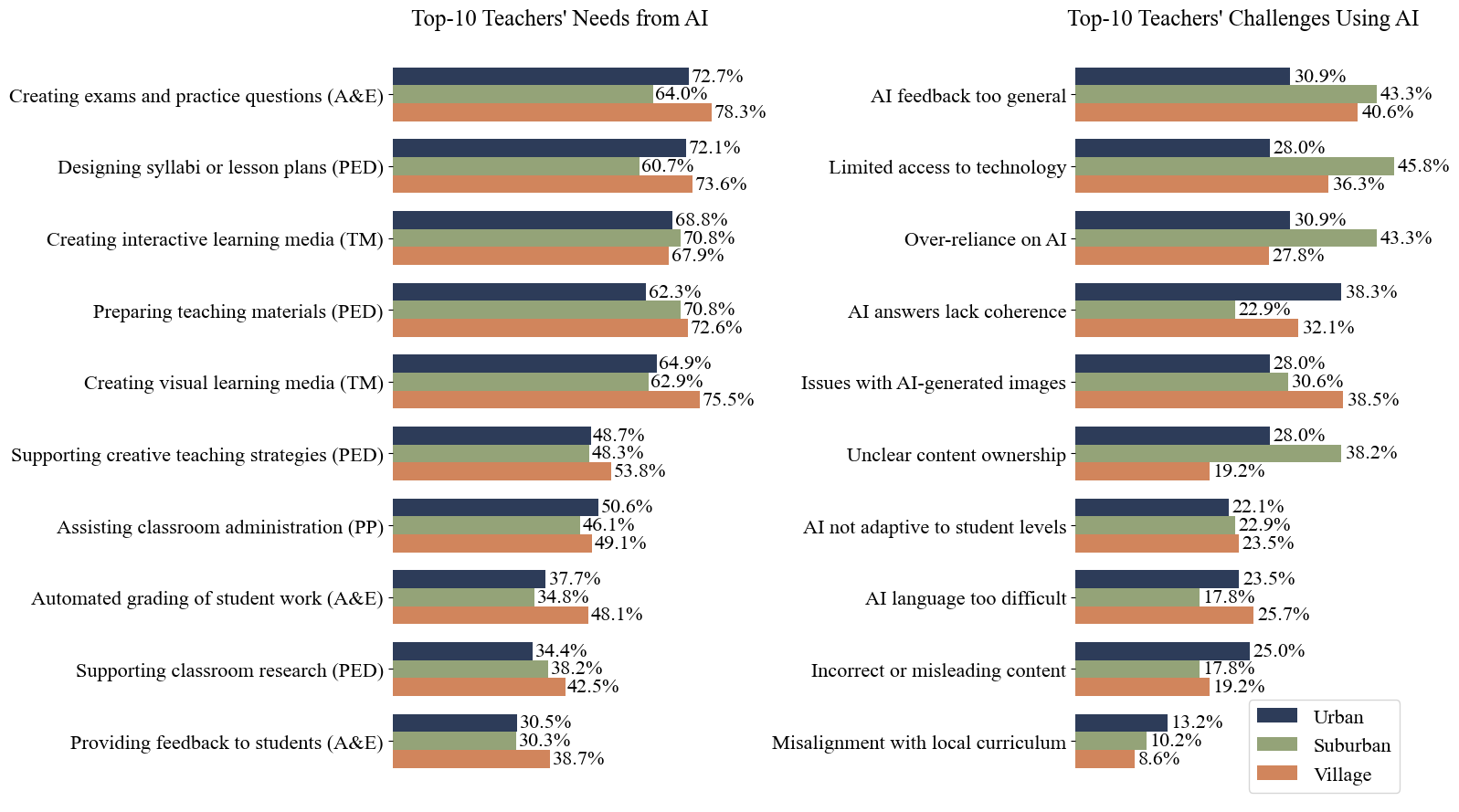}
  \caption{Top‑10 teachers’ needs (left) and top‑10 challenges teachers face (right) when using AI, categorized across school settings (urban, suburban, and village).}
  \label{fig:needs_challenges_of_llmai}
\end{figure}

\section{Result and Analysis}

\subsubsection{Main Result}
The respondents represent a broad span of schooling levels, locations and experience. Most of the respondents teach at Junior High School (41.3\%) and Elementary School (38\%) with Senior High School/Vocational School (26.3\%) is also strongly represented. Schools are distributed across urban (44.1\%), sub-urban (25.4\%) and rural (30.4\%) areas. The experience levels are substantial, with the largest group at 15+ years teaching experience (36.3\%). Furthermore, most respondents report Bachelor degree as their highest qualification (80.4\%) and 67.3\% reported completing teacher professional education program.

Figure~\ref{fig:freq_and_satis_percentage} shows the frequency of use and satisfaction with AI tools across components of effective teaching and learning, disaggregated by school level. Across all levels, teachers most frequently use AI for content knowledge development and teaching media preparation, with over 64\% reporting frequent use, reflecting AI’s value for acquiring, verifying, and presenting subject matter, as well as supporting repetitive and time‑consuming tasks \cite{piyakun-etal-2023-teacher-worklife}. In contrast, curriculum development is the least frequent use case, as it is not a daily activity. Usage patterns are largely consistent across educational levels. Teachers also report generally positive satisfaction with AI, with 66\% indicating satisfaction across components. Teaching media is perceived as the most supported area, while professional practice is the least supported, consistent with its lower usage. Dissatisfaction remains low at around 3\%, while moderate satisfaction is reported by approximately 28\%, indicating room for further improvement.\footnote{A Wilcoxon signed rank test shows statistically significant differences between frequency and satisfaction across all components except content knowledge development ($p < 0.05$).}

\subsubsection{Detailed Results by Teacher Category}

Table~\ref{tab:assessment_detail} summarizes differences in AI tool use across teaching components by teacher characteristics, reporting percentage deviations from component level frequency scores computed from normalized responses to often, sometimes, and rarely. Clear patterns emerge across groups. Teachers in arts, sports, and other subjects report the highest AI use, while language teachers show the lowest usage, followed by science, social science, and general subject teachers. Elementary school teachers use AI more frequently than those at higher levels, whereas teachers with more than 15 years of experience report lower usage. Regionally, teachers in Eastern Indonesia show consistently higher AI use across all components, particularly in pedagogy, curriculum development, and professional practice, each exceeding the overall average by more than 10 percent, indicating strong demand for AI support despite lower infrastructure levels and aligning with prior findings \cite{al-kautsar-etal-2025-indonesians}.

\subsubsection{Types of AI Tools Used by Teachers}
We further analyze the types of AI tools teachers use to support different components of teaching and learning, covering six categories: text based large language models, machine translation, text to speech and speech to text, grammar checkers, image generative AI, and video generative AI. Across all components, text based LLM tools are the most widely used, particularly for content knowledge development, where teachers rely on them for instructional content creation and knowledge acquisition. Machine translation and image generative AI are the next most frequently used tools. Image and video generative AI are used primarily for teaching media, supporting the creation of visual and multimedia materials that enrich instructional content and enhance students’ understanding.

\subsubsection{Percieved Areas of Needs for AI Support}
Figure~\ref{fig:needs_challenges_of_llmai} shows teachers' perceived priorities for AI support across professional tasks. The most frequently selected needs are related to instructional preparation and assessment. Nearly three-quarters of respondents indicated the needs for AI to create assessment items (72.2\%) followed by syllabus design and lesson plan (69.6\%), interactive teaching media (69.1\%), visual teaching materials (67.6\%) and text-based instructional content (67.6\%).  Administrative and pedagogical support also emerged as the next priorities. Around half of the teachers (49.0\%) selected generating teaching method ideas and assisting with administrative tasks (49.0\%). Assessment-related support beyond item generation such as checking student answers (40.1\%) and supporting classroom action research (37.8\%) was moderately needed. 

\subsubsection{Challenges Experienced When Using AI}

Teachers report several challenges when using AI in learning. The most common issue is overly generic feedback (37\%), followed by limited access to technology and infrastructure (35.1\%) and concerns about over reliance on AI (33.1\%). Other notable challenges include unclear AI responses, problems with AI generated images (31.8\%), and uncertainty around intellectual property (27.9\%), while issues such as language clarity, factual inaccuracies, and curriculum misalignment are less prevalent. Overall, while teachers recognize the benefits of AI, they continue to face challenges related to reliability, contextual relevance, infrastructure, and governance.

\section{Conclusion}

In conclusion, these findings point to clear priorities for future AI in education research in Indonesia. The prevalence of overly generic feedback (37\%) and unclear responses highlights the need for AI systems that are more context aware and pedagogically grounded. Infrastructure constraints (35.1\%) underscore the importance of lightweight and offline tolerant designs suited to uneven connectivity, while concerns about over reliance on AI (33.1\%) call for AI tools that support appropriate human–AI interaction when used to assist learning in classroom settings.

\begin{credits}
\subsubsection{\ackname} This research was supported by the 2023 SEAMEO-Australia Education Links Award, jointly organized by the Southeast Asian Ministers of Education Organization (SEAMEO) Secretariat and the Australian Government Department of Education.

\subsubsection{\discintname}
The authors have no competing interests to declare that are relevant to the content of this 
\end{credits}



%
%
%
\bibliographystyle{splncs04}
\bibliography{reference}

\end{document}